\documentclass{article}

\usepackage[dblblindworkshop,final]{neurips_2026}
\workshoptitle{SLMs for Agentic Systems, Paris, France, 2026}
\usepackage[utf8]{inputenc}
\usepackage[T1]{fontenc}
\usepackage{hyperref}
\usepackage{url}
\usepackage{booktabs}
\usepackage{multirow}
\usepackage{amsmath,amssymb}
\usepackage{float}
\usepackage{subcaption}
\usepackage[table]{xcolor}
\usepackage{tikz}
\usepackage{pgfplots}
\pgfplotsset{compat=1.18}
\usepackage{algorithm}
\usepackage{algorithmic}

\definecolor{bf16color}{HTML}{2C73D2}
\definecolor{awqcolor}{HTML}{E74C3C}
\definecolor{gptqcolor}{HTML}{2ECC71}

\title{Beyond Accuracy: How Quantization Affects\\Coding Agents}

\author{%
  Areej Mehboob\textsuperscript{1} \quad
  Muhammad Ali Shafique\textsuperscript{2} \\
  \textbf{Muhammad Usman Qadeer}\textsuperscript{1} \quad
  \textbf{Hamza Farooq}\textsuperscript{1} \\[2pt]
  \textsuperscript{1}Traversaal AI \quad
  \textsuperscript{2}Kansas State University 
  \\[2pt]
  \texttt{areej@traversaal.ai} \quad \texttt{alishafique@ksu.edu} \\
  \texttt{mqadeer@traversaal.ai} \quad \texttt{hamza@traversaal.ai}
}

\begin{document}

\maketitle

\begin{abstract}
Agents are increasingly used for coding tasks, but the inference cost of large language models (LLMs) limits where they can run. Quantization reduces model size, but its effects on coding agents that execute and revise programs remain underexplored. We compare BF16 with 4-bit AWQ and GPTQ for Qwen2.5-Coder-Instruct (7B, 14B, and 32B) and Devstral-Small-2507 (24B) on OJBench coding problems across three seeds. We detect no statistically significant accuracy loss for any of the coding agents. However, final accuracy can conceal substantial behavioral changes: AWQ reduces Qwen-7B one-shot solves from 17 to 0 while increasing loop-based solves from 10 to 25. All Qwen-32B successes involve tool use, whereas Devstral rarely makes a valid executor call. Failure profiles also differ: Qwen-32B produces far more compilation errors than the smaller Qwen models, rising slightly under quantization, while Devstral's failures shift significantly from runtime errors toward wrong answers and timeouts. These findings indicate that quantization robustness in coding agents should be assessed through solution trajectories, tool use, and failure modes, in addition to final accuracy.
\end{abstract}

\section{Introduction}

Large Language Models (LLMs) \citep{jaech2024openai, team2025kimi} are increasingly used in agentic coding systems that go beyond single-pass text generation. These agents can invoke execution tools, inspect program outputs, and revise their solutions over multiple steps \citep{hong2025data, guo2024ds, lai2023ds, chen2024teaching}. This interaction changes how coding capability should be evaluated: a model that produces an incorrect program initially may still reach a correct solution after execution feedback. Recent benchmarks such as OJBench \citep{wang2025ojbenchcompetitionlevelcode} provide challenging settings for studying this behavior. OJBench contains competition-level programming problems and places greater emphasis on reasoning and code correctness than typical interview-style benchmarks.

Efficient deployment of these agents remains challenging because repeated LLM inference and tool interaction increase computational and memory costs. Quantization is a widely used approach for reducing these costs \citep{frantar2023gptqaccurateposttrainingquantization, xiao2023smoothquant, liu2024intactkv}. In particular, weight-only methods such as AWQ and GPTQ can substantially reduce model storage while retaining high-precision activations. However, reasoning performance can be sensitive to low-bit quantization, especially on difficult tasks where small perturbations can affect multi-step generation \citep{liu2025quantizationhurtsreasoningempirical}. Most existing studies evaluate this effect using single-pass model outputs. Coding agents introduce a different setting because quantization can influence not only the generated program, but also tool invocation, iterative refinement, and the final failure mode.

This distinction raises an important question: does similar final accuracy imply that a quantized coding agent behaves similarly to its full-precision counterpart? Our results suggest that it does not. Quantization can preserve final accuracy while changing solve trajectories, tool-use patterns, and terminal failure modes. These effects are hidden when coding agents are evaluated only by their final score. In this work, we present, to the best of our knowledge, the first empirical study of 4-bit weight-only quantization in execution-feedback coding agents. We evaluate BF16, AWQ, and GPTQ versions of Qwen2.5-Coder-Instruct \citep{hui2024qwen25codertechnicalreport} at 7B, 14B, and 32B scales, together with Devstral-Small-2507 \citep{mistral2025devstral}. All models use vLLM with an OpenAI-compatible tool schema and a Python execution tool and are evaluated over three random seeds on OJBench. We investigate three research questions:

RQ1: How robust is end-to-end coding-agent performance to 4-bit quantization?

RQ2: Does end-to-end accuracy conceal changes in coding-agent solution trajectories?

RQ3: Does quantization change how coding agents fail, not just how often?

\begin{table}[t]
\centering
\caption{Performance of Qwen-2.5-Coder and Devstral-Small-2507 agents (vLLM, OpenAI-compatible tool schema) on OJBench, reported as mean$_{\pm\mathrm{std}}$ over 3 seeds. Darker shading marks the best overall accuracy per model family, lighter the second best. Overall accuracy is over the Easy and Medium splits (115 problems); the Hard split is excluded because all models scored 0\%.}
\label{tab:quantized-agents-results}
\small
\begin{tabular}{lcccc}
\toprule
\textbf{Model} & \textbf{Precision} & \textbf{Easy (\%)} & \textbf{Medium (\%)} & \textbf{Overall (\%)}\\
\midrule
\multirow{3}{*}{Qwen-2.5-Coder-7B}
 & BF16 & $21.29_{\pm1.6}$ & $1.69_{\pm0.7}$ & \cellcolor{green!35}$7.83_{\pm0.9}$\\
 & AWQ  & $20.37_{\pm1.6}$ & $1.27_{\pm0.0}$ & \cellcolor{green!15}$7.25_{\pm0.5}$\\
 & GPTQ & $18.52_{\pm3.2}$ & $1.27_{\pm0.0}$ & $6.67_{\pm1.0}$\\
\midrule
\multirow{3}{*}{Qwen-2.5-Coder-14B}
 & BF16 & $30.56_{\pm2.8}$ & $2.53_{\pm1.3}$ & \cellcolor{green!35}$11.30_{\pm0.0}$\\
 & AWQ  & $26.85_{\pm1.6}$ & $1.27_{\pm0.0}$ & $9.28_{\pm0.5}$\\
 & GPTQ & $26.85_{\pm8.9}$ & $2.11_{\pm0.7}$ & \cellcolor{green!15}$9.86_{\pm2.7}$\\
\midrule
\multirow{3}{*}{Qwen-2.5-Coder-32B}
 & BF16 & $32.41_{\pm1.6}$ & $5.48_{\pm3.2}$ & \cellcolor{green!35}$13.91_{\pm2.3}$\\
 & AWQ  & $25.93_{\pm5.8}$ & $3.80_{\pm0.0}$ & $10.72_{\pm1.8}$\\
 & GPTQ & $30.56_{\pm0.0}$ & $2.11_{\pm0.7}$ & \cellcolor{green!15}$11.01_{\pm0.5}$\\
\midrule
\multirow{3}{*}{Devstral-Small-2507}
 & BF16 & $11.11_{\pm2.8}$ & $1.69_{\pm0.7}$ & $4.64_{\pm1.3}$\\
 & AWQ  & $15.74_{\pm8.5}$ & $1.69_{\pm0.7}$ & \cellcolor{green!35}$6.09_{\pm2.6}$\\
 & GPTQ & $13.89_{\pm2.8}$ & $1.69_{\pm0.7}$ & \cellcolor{green!15}$5.51_{\pm1.3}$\\
\bottomrule
\end{tabular}
\end{table}



\section{Related Work}
Recent work on coding agents extends single-pass code generation by incorporating program execution and iterative revision. Studies on self-debugging show that LLMs can use execution feedback to identify and correct errors in generated programs \citep{chen2024teaching}. Existing evaluations cover real-world data-science tasks through DSBC \citep{kadiyala2025dsbcdatascience}, automated program repair \citep{10968728}, interactive application environments such as AppWorld \citep{trivedi2024appworldcontrollableworldapps}, multilingual repository-level tasks through SWE-PolyBench \citep{rashid2025swepolybenchmultilanguagebenchmarkrepository}, and competition-level programming through OJBench \citep{wang2025ojbenchcompetitionlevelcode}.
AWQ \citep{lin2024awqactivationawareweightquantization} and GPTQ \citep{frantar2023gptqaccurateposttrainingquantization} are widely used methods for 4-bit weight-only quantization. Prior single-pass studies indicate that quantization generally preserves reasoning performance, although smaller models and more difficult tasks experience greater degradation \citep{liu2025quantizationhurtsreasoningempirical}. However, these studies do not examine how quantization affects coding agents that can execute and revise their outputs. We address this gap by asking whether the execution loop absorbs quantization-induced errors, and whether the size--robustness ordering observed in single-pass settings still holds when the agent can run its code.

\section{Experimental Setup}

This section describes the setup used to evaluate BF16 and AWQ/GPTQ 4-bit quantized models on competitive coding tasks within an execution-based agentic workflow.

\subsection{Dataset: OJBench}
We evaluate on OJBench \citep{wang2025ojbenchcompetitionlevelcode}, a benchmark of 232 competition-level problems from China's National Olympiad in Informatics (NOI) and the International Collegiate Programming Contest (ICPC). OJBench supports solutions in both Python and C++; we use Python only, so all solutions are written, executed, and judged in Python. We use all 115 Easy and Medium problems because all model--precision configurations scored 0\% on Hard. Evaluation across three seeds yields 345 problem--seed observations per configuration.

\subsection{Evaluation Setup}

\textbf{Models and Quantization} We evaluate Qwen2.5-Coder-Instruct \citep{hui2024qwen25codertechnicalreport} at 7B, 14B, and 32B, spanning scales at which size may interact with quantization, plus the agent-oriented 24B \textit{Devstral-Small-2507} \citep{mistral2025devstral} as an out-of-family comparison. Each model is run at BF16 and as W4A16 variants produced with AWQ \citep{lin2024awqactivationawareweightquantization} and GPTQ \citep{frantar2023gptqaccurateposttrainingquantization} via \textit{llmcompressor} \citep{llmcompressor2024} (256 calibration samples, max sequence length 1,024).

\textbf{Agent Framework.} Each model is served with vLLM in a multi-turn execution-feedback loop using an OpenAI-compatible tool schema, with tool calls parsed from the generated text. The agent has one tool, \texttt{python\_executor(code, stdin)}. It writes code, tests it on the problem's examples, reads the feedback, and revises for up to five tool calls per problem. All runs use nucleus sampling (temperature $0.6$, top-$p$ $0.9$), $32{,}768$-token context, and three seeds. Full prompt and loop in Appendices~\ref{appendix:prompt} and~\ref{appendix:algorithm}.

\textbf{Evaluation Metrics.} Accuracy is the proportion of final programs passing all official OJBench test cases. Each failed attempt is assigned one terminal category: Invalid Return (IR; runtime exception), Wrong Answer (WA), Time Limit Exceeded (TLE), Compilation Error (CE), or Memory Limit Exceeded (MLE). We also record tool calls per problem, distinguishing \emph{one-shot} solves (no tool calls) from \emph{loop-based} solves (at least one execution round). Because every precision is evaluated on the same problems and seeds, outcomes are paired; we apply an exact two-sided McNemar test to the 345 paired observations, with $b$ and $c$ the discordant pairs solved by only the first or only the second precision, and $p<0.05$ as the threshold. For the behavioral measures, we use Fisher's exact test for solve paths and tool-use rates, and a chi-square test for failure distributions. Because each analysis involves eight BF16-versus-quantized comparisons, we also report Holm-adjusted $p$-values. Because the three seeds of a problem are correlated, we additionally treat each problem as one unit: we report 95\% confidence intervals from a cluster bootstrap over the 115 problems (20{,}000 resamples, keeping all three seeds of a problem together) and $p$-values from a permutation test that swaps precision labels within each problem.

\begin{figure*}[t]
    \centering
    \begin{subfigure}[t]{0.48\textwidth}
        \centering
        \resizebox{\textwidth}{!}{%
        \begin{tikzpicture}
            \begin{axis}[
                width=12cm, height=6.5cm, ybar=2pt, bar width=10pt, enlarge x limits=0.25,
                ylabel={\textbf{Count}}, ylabel style={font=\normalsize, at={(axis description cs:-0.08,0.5)}},
                symbolic x coords={IR, WA, TLE, CE, MLE}, xtick=data,
                xticklabel style={font=\large, anchor=north, yshift=-3pt},
                ymin=0, ymax=90, ytick={0, 20, 40, 60, 80}, yticklabel style={font=\normalsize},
                ymajorgrids=true, xmajorgrids=false, grid style={dashed, gray!40, line width=0.4pt},
                axis line style={gray!80, line width=0.6pt},
                legend style={at={(0.5,1.02)}, anchor=south, font=\large, draw=gray!50, fill=white, rounded corners=2pt, legend columns=3, column sep=10pt, inner xsep=8pt, inner ysep=5pt},
                nodes near coords, nodes near coords style={font=\small\bfseries, /pgf/number format/fixed, /pgf/number format/precision=1, color=black},
                nodes near coords align={vertical}, every node near coord/.append style={yshift=4pt}, clip=false, axis on top,
            ]
            \addplot[fill=bf16color, draw=bf16color!70!black, line width=0.4pt] coordinates {(IR, 21.7) (WA, 65.0) (TLE, 17.0) (CE, 1.3) (MLE, 0.7)};
            \addplot[fill=awqcolor, draw=awqcolor!70!black, line width=0.4pt] coordinates {(IR, 29.0) (WA, 59.7) (TLE, 15.0) (CE, 2.3) (MLE, 0.7)};
            \addplot[fill=gptqcolor, draw=gptqcolor!70!black, line width=0.4pt] coordinates {(IR, 23.3) (WA, 66.0) (TLE, 15.0) (CE, 2.0) (MLE, 1.0)};
            \legend{BF16, AWQ, GPTQ}
            \end{axis}
        \end{tikzpicture}%
        }
        \caption{Qwen-2.5-Coder-7B}
        \label{fig:qwen7b_verdicts}
    \end{subfigure}
    \hfill
    \begin{subfigure}[t]{0.48\textwidth}
        \centering
        \resizebox{\textwidth}{!}{%
        \begin{tikzpicture}
            \begin{axis}[
                width=12cm, height=6.5cm, ybar=2pt, bar width=10pt, enlarge x limits=0.25,
                ylabel={\textbf{Count}}, ylabel style={font=\normalsize, at={(axis description cs:-0.08,0.5)}},
                symbolic x coords={IR, WA, TLE, CE, MLE}, xtick=data,
                xticklabel style={font=\large, anchor=north, yshift=-3pt},
                ymin=0, ymax=90, ytick={0, 20, 40, 60, 80}, yticklabel style={font=\normalsize},
                ymajorgrids=true, xmajorgrids=false, grid style={dashed, gray!40, line width=0.4pt},
                axis line style={gray!80, line width=0.6pt},
                legend style={at={(0.5,1.02)}, anchor=south, font=\large, draw=gray!50, fill=white, rounded corners=2pt, legend columns=3, column sep=10pt, inner xsep=8pt, inner ysep=5pt},
                nodes near coords, nodes near coords style={font=\small\bfseries, /pgf/number format/fixed, /pgf/number format/precision=1, color=black},
                nodes near coords align={vertical}, every node near coord/.append style={yshift=4pt}, clip=false, axis on top,
            ]
            \addplot[fill=bf16color, draw=bf16color!70!black, line width=0.4pt] coordinates {(IR, 15.3) (WA, 66.3) (TLE, 18.0) (CE, 2.0) (MLE, 0.3)};
            \addplot[fill=awqcolor, draw=awqcolor!70!black, line width=0.4pt] coordinates {(IR, 13.0) (WA, 69.0) (TLE, 17.0) (CE, 4.3) (MLE, 1.0)};
            \addplot[fill=gptqcolor, draw=gptqcolor!70!black, line width=0.4pt] coordinates {(IR, 21.0) (WA, 61.3) (TLE, 18.7) (CE, 2.0) (MLE, 0.7)};
            \legend{BF16, AWQ, GPTQ}
            \end{axis}
        \end{tikzpicture}%
        }
        \caption{Qwen-2.5-Coder-14B}
        \label{fig:qwen14b_verdicts}
    \end{subfigure}
    \par\vspace{0.8em}
    \begin{subfigure}[t]{0.48\textwidth}
        \centering
        \resizebox{\textwidth}{!}{%
        \begin{tikzpicture}
            \begin{axis}[
                width=12cm, height=6.5cm, ybar=2pt, bar width=10pt, enlarge x limits=0.25,
                ylabel={\textbf{Count}}, ylabel style={font=\normalsize, at={(axis description cs:-0.08,0.5)}},
                symbolic x coords={IR, WA, TLE, CE, MLE}, xtick=data,
                xticklabel style={font=\large, anchor=north, yshift=-3pt},
                ymin=0, ymax=90, ytick={0, 20, 40, 60, 80}, yticklabel style={font=\normalsize},
                ymajorgrids=true, xmajorgrids=false, grid style={dashed, gray!40, line width=0.4pt},
                axis line style={gray!80, line width=0.6pt},
                legend style={at={(0.5,1.02)}, anchor=south, font=\large, draw=gray!50, fill=white, rounded corners=2pt, legend columns=3, column sep=10pt, inner xsep=8pt, inner ysep=5pt},
                nodes near coords, nodes near coords style={font=\small\bfseries, /pgf/number format/fixed, /pgf/number format/precision=1, color=black},
                nodes near coords align={vertical}, every node near coord/.append style={yshift=4pt}, clip=false, axis on top,
            ]
            \addplot[fill=bf16color, draw=bf16color!70!black, line width=0.4pt] coordinates {(IR, 6.7) (WA, 63.7) (TLE, 15.0) (CE, 13.3) (MLE, 0.3)};
            \addplot[fill=awqcolor, draw=awqcolor!70!black, line width=0.4pt] coordinates {(IR, 7.7) (WA, 66.7) (TLE, 10.7) (CE, 17.7) (MLE, 0.0)};
            \addplot[fill=gptqcolor, draw=gptqcolor!70!black, line width=0.4pt] coordinates {(IR, 5.7) (WA, 58.7) (TLE, 16.7) (CE, 20.7) (MLE, 0.7)};
            \legend{BF16, AWQ, GPTQ}
            \end{axis}
        \end{tikzpicture}%
        }
        \caption{Qwen-2.5-Coder-32B}
        \label{fig:qwen32b_verdicts}
    \end{subfigure}
    \hfill
    \begin{subfigure}[t]{0.48\textwidth}
        \centering
        \resizebox{\textwidth}{!}{%
        \begin{tikzpicture}
            \begin{axis}[
                width=12cm, height=6.5cm, ybar=2pt, bar width=10pt, enlarge x limits=0.25,
                ylabel={\textbf{Count}}, ylabel style={font=\normalsize, at={(axis description cs:-0.08,0.5)}},
                symbolic x coords={IR, WA, TLE, CE, MLE}, xtick=data,
                xticklabel style={font=\large, anchor=north, yshift=-3pt},
                ymin=0, ymax=90, ytick={0, 20, 40, 60, 80}, yticklabel style={font=\normalsize},
                ymajorgrids=true, xmajorgrids=false, grid style={dashed, gray!40, line width=0.4pt},
                axis line style={gray!80, line width=0.6pt},
                legend style={at={(0.5,1.02)}, anchor=south, font=\large, draw=gray!50, fill=white, rounded corners=2pt, legend columns=3, column sep=10pt, inner xsep=8pt, inner ysep=5pt},
                nodes near coords, nodes near coords style={font=\small\bfseries, /pgf/number format/fixed, /pgf/number format/precision=1, color=black},
                nodes near coords align={vertical}, every node near coord/.append style={yshift=4pt}, clip=false, axis on top,
            ]
            \addplot[fill=bf16color, draw=bf16color!70!black, line width=0.4pt] coordinates {(IR, 53.3) (WA, 44.7) (TLE, 7.0) (CE, 4.7) (MLE, 0.0)};
            \addplot[fill=awqcolor, draw=awqcolor!70!black, line width=0.4pt] coordinates {(IR, 37.3) (WA, 51.0) (TLE, 10.0) (CE, 9.0) (MLE, 0.7)};
            \addplot[fill=gptqcolor, draw=gptqcolor!70!black, line width=0.4pt] coordinates {(IR, 39.3) (WA, 51.7) (TLE, 14.3) (CE, 3.3) (MLE, 0.0)};
            \legend{BF16, AWQ, GPTQ}
            \end{axis}
        \end{tikzpicture}%
        }
        \caption{Devstral-Small-2507}
        \label{fig:devstral_verdicts}
    \end{subfigure}
    \vspace{2pt}
    \caption{Distribution of failure categories across BF16, AWQ, and GPTQ for each agent. Categories: IR (Invalid Return), WA (Wrong Answer), TLE (Time Limit Exceeded), CE (Compilation Error), MLE (Memory Limit Exceeded). Counts are per-seed averages over the Easy+Medium splits (115 problems, 3 seeds).}
    \label{fig:agent_error_analysis}
\end{figure*}
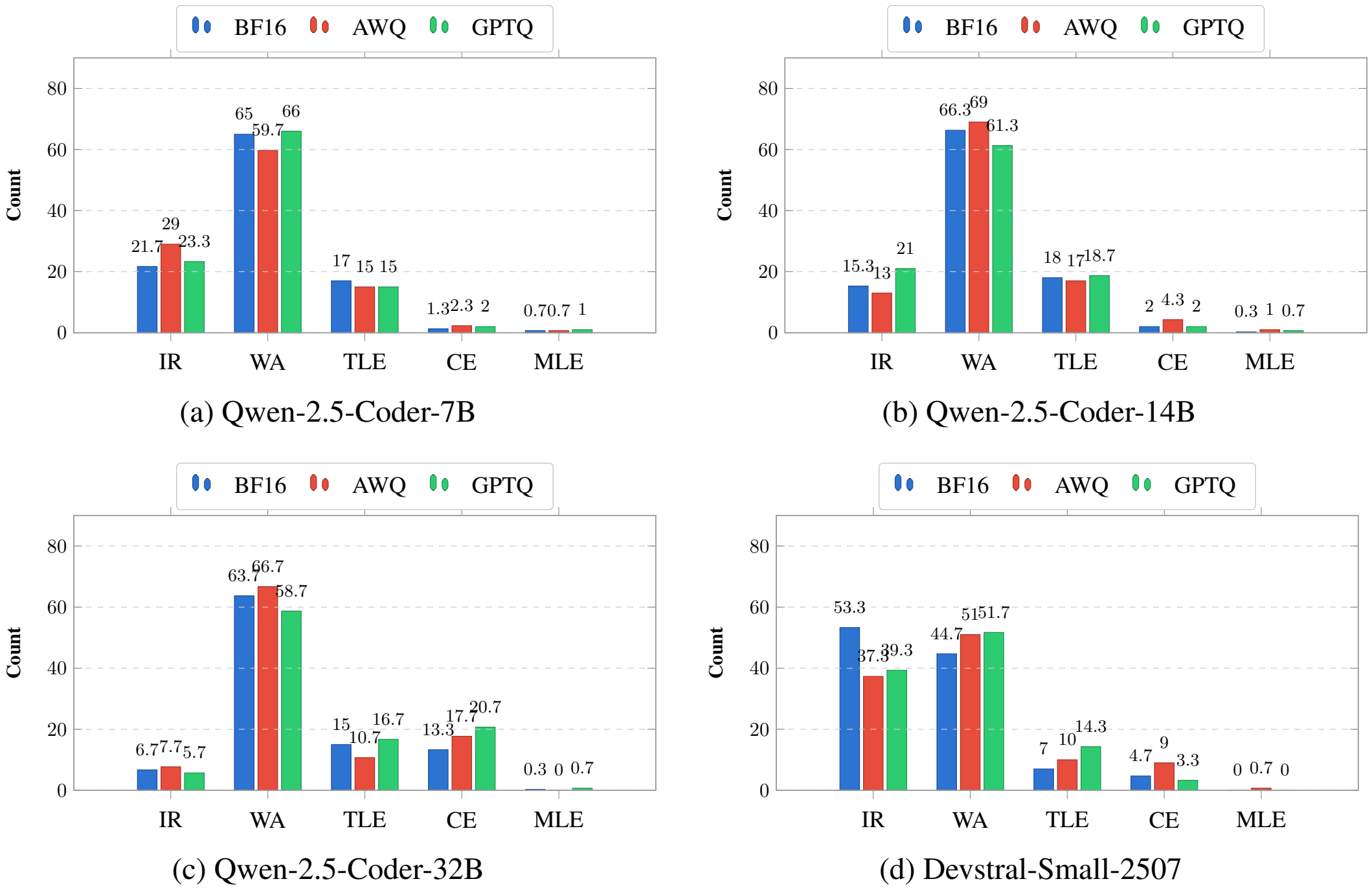

\section{Empirical Findings}

We investigate how weight-only quantization affects Python coding performance on OJBench for execution-feedback agents across BF16, AWQ, and GPTQ 4-bit precisions. Table~\ref{tab:quantized-agents-results} reports accuracy, Figure~\ref{fig:agent_error_analysis} the failure-mode distribution, and Appendix~\ref{appendix:stats} the paired tests and tool-use statistics.

\textbf{RQ1: How robust is end-to-end coding-agent performance to 4-bit quantization?}

We evaluate whether replacing BF16 weights with W4A16 AWQ or GPTQ changes final coding accuracy in execution-feedback agents. We compare accuracy across Qwen-2.5-Coder scales and Devstral-Small-2507, together with paired statistical tests in Table~\ref{tab:mcnemar}. This establishes whether quantization-induced perturbations remain visible after the complete agentic execution process and whether the effect is consistent across quantizers and model configurations. Because we run eight BF16-versus-quantized comparisons, we apply a Holm correction, and because the three seeds of a problem are correlated, we also repeat the analysis with each problem as one unit. Under both analyses, no comparison shows a statistically significant accuracy loss. Qwen-32B shows the largest drop, about 3 points under both quantizers, but it does not reach significance after correction ($p_{\mathrm{Holm}}=0.21$ for AWQ) or at the problem level ($p=0.051$).

\emph{Finding: Quantization has limited effect on final accuracy for all agents. Qwen-2.5-Coder-32B shows the largest drop, but it does not reach statistical significance.}

\textbf{RQ2: Does end-to-end accuracy conceal changes in coding-agent solution trajectories?}

We analyze successful trajectories using one-shot solves, execution-loop solves, tool-use rate, and mean executor calls (Tables~\ref{tab:tool-use} and~\ref{tab:fisher}). Across precisions, Qwen-7B shows the clearest behavioral shift: one-shot solves fall from 17 under BF16 to 0 under AWQ and 3 under GPTQ (AWQ: 17 $\rightarrow$ 0; GPTQ: 17 $\rightarrow$ 3), while loop-based solves rise from 10 under BF16 to 25 under AWQ and 20 under GPTQ (AWQ: 10 $\rightarrow$ 25; GPTQ: 10 $\rightarrow$ 20). This shift remains significant when each problem is treated as one unit, and it also holds on problems solved by both precisions. Devstral moves in the opposite direction, with valid tool use falling from 8.7\% under BF16 to 1.2\% under GPTQ (GPTQ: 8.7\% $\rightarrow$ 1.2\%), where all 19 successful solves are one-shot. Malformed tool calls are rare for the Qwen models, but most of Devstral's tool calls do not follow the tool schema, which keeps its recorded tool use low; quantization still reduces how often it tries the tool (AWQ: 42.0\% $\rightarrow$ 27.0\%; GPTQ: 42.0\% $\rightarrow$ 21.4\%). The Qwen models also differ from each other at BF16: Qwen-32B uses tools on 71.3\% of problems and produces no one-shot solves, whereas Qwen-14B uses tools on 31.6\% of problems with a near-balanced split of 21 one-shot and 18 loop-based solves. These results show both model-specific solve strategies and different levels of behavioral stability under quantization.

\emph{Finding: Models exhibit distinct tool-use regimes, but quantization does not preserve them uniformly. Qwen-32B remains loop-dependent, Qwen-14B stays relatively balanced, Qwen-7B shifts toward loop-based solving, and Devstral becomes almost entirely one-shot under GPTQ.}

\textbf{RQ3: Does quantization change how coding agents fail, not just how often?}

We compare terminal failure categories (IR /WA /TLE /CE /MLE) across precision settings, and test whether each failure distribution differs from BF16 using a chi-square test with Holm correction (Table~\ref{tab:error-analysis}). For Qwen-7B, AWQ shifts failures slightly away from Wrong Answer and toward Invalid Return (WA: $65.0 \rightarrow 59.7$; IR: $21.7 \rightarrow 29.0$ per seed). This suggests that quantization can change the type of terminal failure even when final accuracy changes little. The pattern differs for larger models. Qwen-32B exhibits substantially more Compilation Errors than the smaller Qwen models, with CE rising slightly under quantization. Devstral follows another pattern, with fewer Invalid Returns but more Wrong Answers and Time Limit Exceeded cases after quantization. This is the clearest change, and it is statistically significant under both quantizers.

\emph{Finding: Quantization can reshape failure profiles differently across models rather than producing one universal failure pattern.}

\section{Conclusion}

We evaluated 4-bit weight-only quantization in execution-feedback coding agents on OJBench, comparing BF16, AWQ, and GPTQ across Qwen2.5-Coder (7B, 14B, 32B) and Devstral-Small-2507. Quantization reduces checkpoint storage by about 3.4$\times$ and causes no statistically significant accuracy loss for any agent after correcting for multiple comparisons, although Qwen-32B shows a consistent nominal drop. More importantly, similar final accuracy can hide large behavioral shifts: Qwen-7B moves strongly from one-shot to loop-based solving after quantization, while Qwen-32B remains fully loop-dependent. Quantization also changes terminal failure profiles differently across models. These results show that quantized coding agents should be evaluated through accuracy, solve trajectories, tool use, and failure modes rather than final accuracy alone.

\textbf{Limitations and future work.} Our study focuses on one competition-level benchmark, one execution-feedback scaffold, and W4A16 quantization, with Python as the solution language. Because OJBench is challenging, absolute solve counts are modest, so we emphasize effects that hold at the problem level. Extending this analysis to other benchmarks, languages, and bit-widths, and to end-to-end efficiency measures such as latency and memory, is a natural next step. Within these limits, 4-bit quantization appears practical for execution-feedback coding agents, but its reliability should be evaluated within the complete agentic workflow.

\section*{Acknowledgments}

The authors used Claude Opus 5.5 for language and formatting assistance. All content was reviewed by the authors, who remain responsible for the final manuscript.

\bibliographystyle{plainnat}
\bibliography{conference}

\appendix
\raggedbottom
\section{Appendix}

\subsection{Error Analysis}
\label{appendix:errors}

\begin{table}[H]
\centering
\caption{Distribution of Failure Types in Agent Submissions. IR: Invalid Return, WA: Wrong Answer, TLE: Time Limit Exceeded, CE: Compilation Error, MLE: Memory Limit Exceeded. Counts are per-seed averages over the Easy+Medium splits (115 problems, 3 seeds). $p$ is a chi-square test of whether the failure distribution differs from BF16, computed on totals over the three seeds; MLE is excluded from the test because its counts are very small. $p_{\mathrm{Holm}}$ applies a Holm correction across the eight comparisons.}
\label{tab:error-analysis}
\small
\begin{tabular}{lcccccccc}
\toprule
\textbf{Model} & \textbf{Precision} & \textbf{IR} & \textbf{WA} & \textbf{TLE} & \textbf{CE} & \textbf{MLE} & $p$ & $p_{\mathrm{Holm}}$ \\
\midrule
\multirow{3}{*}{Qwen-2.5-Coder-7B}
 & BF16 & 21.7 & 65.0 & 17.0 & 1.3 & 0.7 & -- & -- \\
 & AWQ  & 29.0 & 59.7 & 15.0 & 2.3 & 0.7 & 0.17 & 0.84 \\
 & GPTQ & 23.3 & 66.0 & 15.0 & 2.0 & 1.0 & 0.81 & 1.00 \\
\midrule
\multirow{3}{*}{Qwen-2.5-Coder-14B}
 & BF16 & 15.3 & 66.3 & 18.0 & 2.0 & 0.3 & -- & -- \\
 & AWQ  & 13.0 & 69.0 & 17.0 & 4.3 & 1.0 & 0.34 & 1.00 \\
 & GPTQ & 21.0 & 61.3 & 18.7 & 2.0 & 0.7 & 0.35 & 1.00 \\
\midrule
\multirow{3}{*}{Qwen-2.5-Coder-32B}
 & BF16 & 6.7 & 63.7 & 15.0 & 13.3 & 0.3 & -- & -- \\
 & AWQ  & 7.7 & 66.7 & 10.7 & 17.7 & 0.0 & 0.24 & 0.97 \\
 & GPTQ & 5.7 & 58.7 & 16.7 & 20.7 & 0.7 & 0.13 & 0.75 \\
\midrule
\multirow{3}{*}{Devstral-Small-2507}
 & BF16 & 53.3 & 44.7 & 7.0  & 4.7 & 0.0 & -- & -- \\
 & AWQ  & 37.3 & 51.0 & 10.0 & 9.0 & 0.7 & \textbf{0.002} & \textbf{0.011} \\
 & GPTQ & 39.3 & 51.7 & 14.3 & 3.3 & 0.0 & \textbf{0.001} & \textbf{0.009} \\
\bottomrule
\end{tabular}
\end{table}

\subsection{Statistical Tests and Solve Path}
\label{appendix:stats}

\begin{table}[H]
\centering
\caption{Exact two-sided McNemar tests, BF16 versus quantized, over paired per-problem outcomes (same problem, same seed; 345 pairs per comparison). $b$ = problems solved only by BF16; $c$ = problems solved only by the quantized model. $p_{\mathrm{Holm}}$ applies a Holm correction across the eight comparisons. The 32B/AWQ comparison is significant before correction ($p=0.027$, bold) but not after it; no comparison is significant after correction. The last three columns treat each problem as one unit: $\Delta$ is BF16 minus quantized accuracy (points), the 95\% CI comes from a cluster bootstrap over the 115 problems, and $p_{\mathrm{prob}}$ from a within-problem permutation test. The problem-level analysis gives the same conclusion: no comparison is significant at $\alpha=0.05$, although the 32B/AWQ result is borderline (its CI just excludes zero, $p=0.051$).}
\label{tab:mcnemar}
\footnotesize
\setlength{\tabcolsep}{4pt}
\begin{tabular}{llccccccc}
\toprule
 & & \multicolumn{4}{c}{\textbf{Paired (345 pairs)}} & \multicolumn{3}{c}{\textbf{Problem level (115 problems)}} \\
\cmidrule(lr){3-6}\cmidrule(lr){7-9}
\textbf{Model} & \textbf{Scheme} & $b$ & $c$ & $p$ & $p_{\mathrm{Holm}}$ & $\Delta$ & 95\% CI & $p_{\mathrm{prob}}$ \\
\midrule
\multirow{2}{*}{Qwen-2.5-Coder-7B}  & AWQ  & 8  & 6  & 0.79 & 1.00 & $0.58$ & $[-1.74, 3.19]$ & 0.82 \\
                                    & GPTQ & 8  & 4  & 0.39 & 1.00 & $1.16$ & $[-0.87, 3.19]$ & 0.42 \\
\midrule
\multirow{2}{*}{Qwen-2.5-Coder-14B} & AWQ  & 9  & 2  & 0.065 & 0.46 & $2.03$ & $[0.00, 4.06]$ & 0.094 \\
                                    & GPTQ & 9  & 4  & 0.27 & 1.00 & $1.45$ & $[-0.58, 3.48]$ & 0.27 \\
\midrule
\multirow{2}{*}{Qwen-2.5-Coder-32B} & AWQ  & 16 & 5  & \textbf{0.027} & 0.21 & $3.19$ & $[0.58, 6.09]$ & 0.051 \\
                                    & GPTQ & 18 & 8  & 0.076 & 0.46 & $2.90$ & $[-0.29, 6.38]$ & 0.13 \\
\midrule
\multirow{2}{*}{Devstral-Small-2507}& AWQ  & 7  & 12 & 0.36 & 1.00 & $-1.45$ & $[-3.48, 0.58]$ & 0.27 \\
                                    & GPTQ & 8  & 11 & 0.65 & 1.00 & $-0.87$ & $[-3.48, 1.45]$ & 0.65 \\
\bottomrule
\end{tabular}
\end{table}

\begin{table}[H]
\centering
\caption{Tool use and solve path per agent, totals over 3 seeds (345 attempts). ``Used tool'' is the share of problems with at least one valid executor call. ``Tried tool'' also counts attempts in which the model issued a tool call that did not follow the provided tool schema (e.g., broken JSON or the tool written as a Python function inside its code). ``Malformed'' is the share of all tool calls that did not follow the schema (mean over seeds). $\dagger$ marks tried-tool rates that differ significantly from BF16 after Holm correction. ``One-shot'' solves used zero tool calls; ``loop'' solves followed at least one execute-and-check round.}
\label{tab:tool-use}
\footnotesize
\setlength{\tabcolsep}{4pt}
\begin{tabular}{lccccccc}
\toprule
 & & \textbf{Used} & \textbf{Tried} & \textbf{Malformed} & \textbf{Mean} & \textbf{One-shot} & \textbf{Loop} \\
\textbf{Model} & \textbf{Precision} & \textbf{tool (\%)} & \textbf{tool (\%)} & \textbf{(\%)} & \textbf{calls} & \textbf{solves} & \textbf{solves} \\
\midrule
\multirow{3}{*}{Qwen-2.5-Coder-7B}
 & BF16 & 66.7 & 70.1 & 3.8 & 2.26 & 17 & 10 \\
 & AWQ  & 73.9 & 82.9$^\dagger$ & 5.9 & 2.29 & 0  & 25 \\
 & GPTQ & 69.0 & 71.9 & 3.4 & 1.90 & 3  & 20 \\
\midrule
\multirow{3}{*}{Qwen-2.5-Coder-14B}
 & BF16 & 31.6 & 31.6 & 0.0 & 1.21 & 21 & 18 \\
 & AWQ  & 26.7 & 26.7 & 0.0 & 1.01 & 17 & 15 \\
 & GPTQ & 34.8 & 35.9 & 0.8 & 1.55 & 17 & 17 \\
\midrule
\multirow{3}{*}{Qwen-2.5-Coder-32B}
 & BF16 & 71.3 & 71.6 & 0.1 & 3.07 & 0 & 48 \\
 & AWQ  & 63.8 & 64.3 & 0.4 & 2.69 & 0 & 37 \\
 & GPTQ & 67.0 & 69.6 & 0.9 & 3.15 & 0 & 38 \\
\midrule
\multirow{3}{*}{Devstral-Small-2507}
 & BF16 & 8.7 & 42.0 & 76.9 & 0.11 & 13 & 3 \\
 & AWQ  & 8.4 & 27.0$^\dagger$ & 69.3 & 0.09 & 17 & 4 \\
 & GPTQ & 1.2 & 21.4$^\dagger$ & 96.5 & 0.01 & 19 & 0 \\
\bottomrule
\end{tabular}
\end{table}

\begin{table}[H]
\centering
\caption{Fisher's exact tests on the behavioral measures in Table~\ref{tab:tool-use}, BF16 versus quantized. \emph{Solve path} tests whether the one-shot/loop split among solved problems differs from BF16; \emph{tool use} tests whether the share of problems with at least one executor call differs from BF16. $p_{\mathrm{Holm}}$ applies a Holm correction across the eight comparisons within each measure. Significant results after correction are in bold. The tests treat runs as independent, so $p$-values are approximate. At the problem level, the Qwen-7B solve-path shift is significant under AWQ ($p=0.005$) and GPTQ ($p=0.022$). On problems solved by both precisions, every BF16 one-shot solve becomes a loop solve under AWQ (8 of 8 problems, $p=0.008$), and 7 of 8 under GPTQ ($p=0.070$), which does not reach significance on its own.}
\label{tab:fisher}
\small
\begin{tabular}{llcccc}
\toprule
 & & \multicolumn{2}{c}{\textbf{Solve path}} & \multicolumn{2}{c}{\textbf{Tool use}} \\
\cmidrule(lr){3-4}\cmidrule(lr){5-6}
\textbf{Model} & \textbf{Scheme} & $p$ & $p_{\mathrm{Holm}}$ & $p$ & $p_{\mathrm{Holm}}$ \\
\midrule
\multirow{2}{*}{Qwen-2.5-Coder-7B}  & AWQ  & $<$0.001 & \textbf{$<$0.001} & 0.045 & 0.29 \\
                                    & GPTQ & $<$0.001 & \textbf{0.003} & 0.57 & 1.00 \\
\midrule
\multirow{2}{*}{Qwen-2.5-Coder-14B} & AWQ  & 1.00 & 1.00 & 0.18 & 0.90 \\
                                    & GPTQ & 0.82 & 1.00 & 0.42 & 1.00 \\
\midrule
\multirow{2}{*}{Qwen-2.5-Coder-32B} & AWQ  & 1.00 & 1.00 & 0.042 & 0.29 \\
                                    & GPTQ & 1.00 & 1.00 & 0.25 & 0.99 \\
\midrule
\multirow{2}{*}{Devstral-Small-2507}& AWQ  & 1.00 & 1.00 & 1.00 & 1.00 \\
                                    & GPTQ & 0.086 & 0.51 & $<$0.001 & \textbf{$<$0.001} \\
\bottomrule
\end{tabular}
\end{table}

\begin{figure}[H]
\centering
\begin{tikzpicture}
\begin{axis}[
    width=0.99\textwidth, height=5.9cm,
    ybar=3pt, bar width=13pt,
    enlarge x limits=0.18,
    ymin=0, ymax=18,
    ylabel={Overall accuracy (\%)},
    ylabel style={font=\small},
    symbolic x coords={Qwen-7B, Qwen-14B, Qwen-32B, Devstral},
    xtick=data,
    xticklabel style={font=\small},
    yticklabel style={font=\small},
    ymajorgrids=true, grid style={dashed, gray!30},
    axis line style={gray!70},
    legend style={at={(0.02,0.98)}, anchor=north west, font=\small, draw=gray!40, legend columns=3, column sep=6pt},
    error bars/y dir=both, error bars/y explicit,
    error bars/error bar style={gray!60!black, line width=0.6pt},
    error bars/error mark options={rotate=90, gray!60!black, mark size=2pt},
    clip=false,
]
\addplot[fill=bf16color, draw=bf16color!70!black] coordinates {
  (Qwen-7B,7.83) +- (0,0.9) (Qwen-14B,11.30) +- (0,0.0) (Qwen-32B,13.91) +- (0,2.3) (Devstral,4.64) +- (0,1.3)};
\addplot[fill=awqcolor, draw=awqcolor!70!black] coordinates {
  (Qwen-7B,7.25) +- (0,0.5) (Qwen-14B,9.28) +- (0,0.5) (Qwen-32B,10.72) +- (0,1.8) (Devstral,6.09) +- (0,2.6)};
\addplot[fill=gptqcolor, draw=gptqcolor!70!black] coordinates {
  (Qwen-7B,6.67) +- (0,1.0) (Qwen-14B,9.86) +- (0,2.7) (Qwen-32B,11.01) +- (0,0.5) (Devstral,5.51) +- (0,1.3)};
\legend{BF16, AWQ 4-bit, GPTQ 4-bit}
\node[font=\small\bfseries, awqcolor!80!black] at (axis cs:Qwen-32B,16.9) {$\ast$ $p=0.027$ (n.s.\ after Holm)};
\end{axis}
\end{tikzpicture}
\caption{Overall agent accuracy on OJBench (Easy+Medium, 115 problems), mean over 3 seeds with one standard deviation. The 32B agent under AWQ is the only comparison that differs from BF16 before correction (paired McNemar, $\ast$); after Holm correction across the eight comparisons, no gap is statistically significant.}
\label{fig:accuracy}
\end{figure}
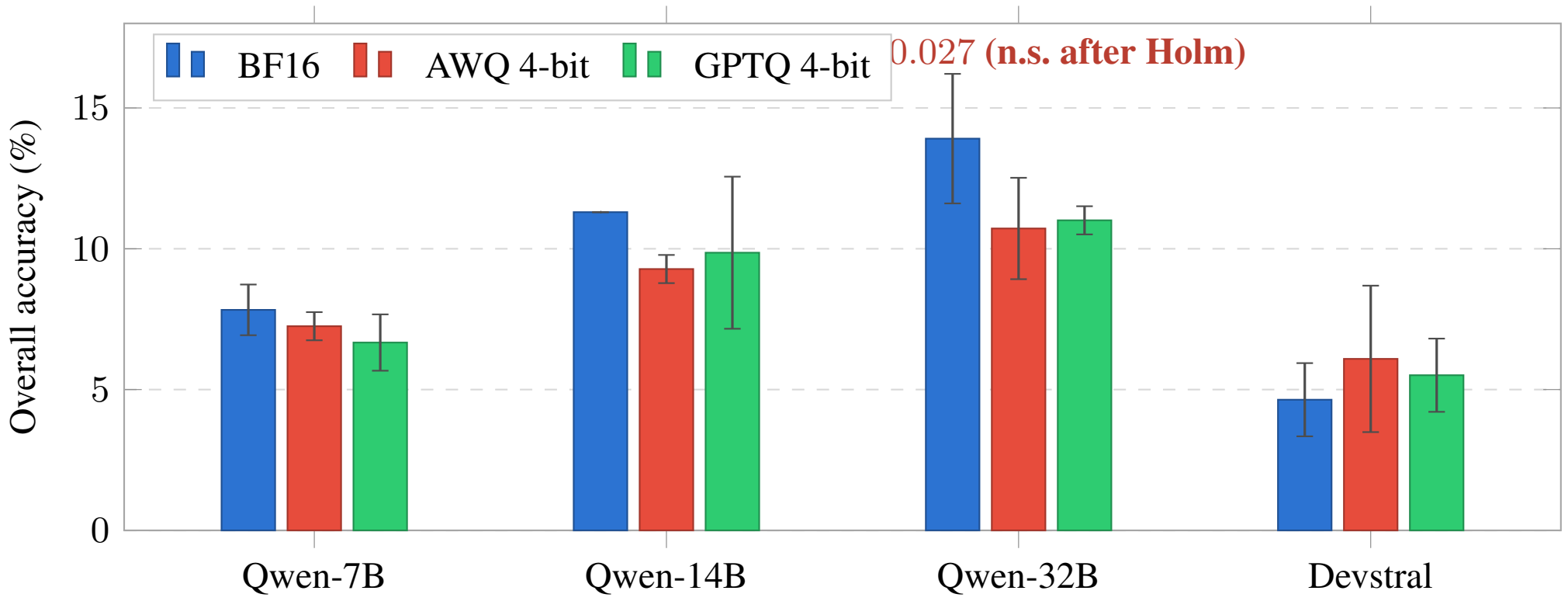

\subsection{Coding Agent Execution Prompt}
\label{appendix:prompt}

The following system prompt was used to instruct the coding agent during problem solving. It guides the agent through a structured write--test--fix workflow with code execution for verification.

\begin{quote}
{\ttfamily
You are a competitive programming assistant. Solve problems step-by-step:\\
1. Read the problem carefully\\
2. Write Python code to solve it\\
3. Create test input from the problem's examples\\
4. Call python\_executor tool to test your code with the test input\\
5. Check if the output matches expected output\\
6. If wrong, fix the code and call python\_executor again\\
7. Once correct, return your final solution in a Python code block\\
Input reading patterns:\\
\hspace*{1.5em}- int(input()) for single number\\
\hspace*{1.5em}- map(int, input().split()) for multiple numbers on one line\\
Always test your code before giving the final answer!
}
\end{quote}

The \texttt{python\_executor(code, stdin)} tool is not described in the prompt text itself. It is supplied to the model through the chat template's OpenAI-compatible tool schema, and tool calls are parsed from the generated text. The tool description reads: \textit{``Execute Python code with optional stdin input. Use this to test your solution before giving the final answer.''} The 30\,s execution timeout and the five-call budget are enforced by the harness and are not disclosed to the model in advance. When the budget is reached, the model receives one additional user message: \textit{``You have reached the maximum number of tool calls. Please provide your final solution in a Python code block now.''}

\subsection{Algorithm}
\label{appendix:algorithm}

Each turn consists of a single model generation. If the generation contains no tool call, it is taken as the final answer. Otherwise every tool call in the generation is executed and its output, truncated to 2{,}000 characters, is appended to the conversation. The tool-call budget $K$ is checked once per turn rather than once per call, so a turn containing several calls can carry the total past $K$; this occurred in roughly 10\% of problem attempts. When the budget is reached, the model is prompted once more to emit its final solution. If the accumulated conversation exceeds the 32{,}768-token context limit, the oldest assistant and tool messages are dropped before generation.

\begin{algorithm}[H]
\caption{Execution-Guided Coding Agent with Tool Feedback}
\footnotesize
\begin{algorithmic}[1]
\REQUIRE Prompt dataset $D$, model $M$, python executor $E$, tool-call budget $K=5$
\ENSURE Final solution and tool-call count for each problem
\FOR{each prompt $x \in D$}
    \STATE messages $\leftarrow$ [system prompt, $x$]
    \STATE tool\_calls $\leftarrow 0$
    \STATE final\_response $\leftarrow \emptyset$
    \FOR{turn $= 1, \ldots, K+1$}
        \STATE Trim oldest assistant and tool messages if context budget is exceeded
        \STATE Generate response $r$ from model $M$ on messages
        \STATE Parse tool calls $T$ from the text of $r$
        \IF{$T$ is empty}
            \STATE final\_response $\leftarrow r$
            \STATE Break loop
        \ENDIF
        \FOR{each call $t \in T$}
            \STATE tool\_calls $\leftarrow$ tool\_calls + 1
            \STATE Execute $t$ using executor $E$ and truncate output to 2{,}000 characters
            \STATE Append $r$ and execution result to messages
        \ENDFOR
        \IF{tool\_calls $\ge K$}
            \STATE Append user message requesting the final solution
            \STATE Generate final\_response from model $M$ on messages
            \STATE Break loop
        \ENDIF
    \ENDFOR
    \IF{final\_response is empty}
        \STATE final\_response $\leftarrow$ last $r$, or error message if no generation succeeded
    \ENDIF
    \STATE Save result including metadata and tool\_calls
\ENDFOR
\end{algorithmic}
\end{algorithm}

\end{document}